\pdfoutput=1

\documentclass[11pt]{article}

\usepackage[final]{acl}
\usepackage{times}
\usepackage{latexsym}
\usepackage[T1]{fontenc}
\usepackage[utf8]{inputenc}
\usepackage{microtype}
\usepackage{inconsolata}

\usepackage{amsmath,amssymb}
\usepackage{booktabs}
\usepackage{enumitem}
\usepackage{multirow}
\usepackage{graphicx}
\usepackage{url}

\title{NBF at SemEval-2025 Task 5: Light-Burst Attention Enhanced System for Multilingual Subject Recommendation}

\author{
Baharul Islam\textsuperscript{*}, 
Nasim Ahmad\textsuperscript{\dag}, 
Ferdous Ahmed Barbhuiya\textsuperscript{*}, 
Kuntal Dey\textsuperscript{*} \\
\textsuperscript{*}IIIT Guwahati \quad 
\textsuperscript{\dag}Dibrugarh University \\
\texttt{\{baharul.islam22b,ferdous,kuntal.dey\}@iiitg.ac.in, nasimahmad4117@gmail.com}
}

\begin{document}
\maketitle

\begin{abstract}
We present our system submission for SemEval 2025 Task 5,  which focuses on cross-lingual subject classification in the English and German academic domains. Our approach leverages bilingual data during training, employing negative sampling and a margin-based retrieval objective. We demonstrate that a dimension-as-token self-attention mechanism designed with significantly reduced internal dimensions can effectively encode sentence embeddings for subject retrieval. In quantitative evaluation, our system achieved an average recall rate of 32.24\% in the general quantitative setting (all subjects), 43.16\% and 31.53\% of the general qualitative evaluation methods with minimal GPU usage, highlighting their competitive performance. Our results demonstrate that our approach is effective in capturing relevant subject information under resource constraints, although there is still room for improvement.
\end{abstract}

\section{Introduction}
Automated subject classification of scholarly articles is of growing importance for digital repositories, allowing more efficient data retrieval, recommendation, and systematic reviews \citep{devlin2019bert}. At \textbf{SemEval-2025 Task 5} \citep{dsouza-EtAl:2025:SemEval2025}, participants must predict \emph{subject codes} for articles in two languages (English and German). 

Previous SemEval tasks on concept or subject classification have used both classical machine learning and deep approaches \citep{agirre2014semeval}. Large pre-trained models (e.g., GPT-based \citep{brown2020language} or other generative approaches) could be powerful but are often computationally heavy. We instead adopt a \emph{dimension-as-token} approach, refining base embeddings from Sentence Transformers \citep{reimers2019sentence} via an ultra-light attention transform. 

By participating in \textbf{SemEval-2025 Task 5}, we discovered that our parameter-efficient, dimension-as-token self-attention approach can effectively handle both English and German data, achieving moderate recall rates of 32. 24\% in the overall quantitative setting (all subjects), 43. 16\% in Case 1 and 31. 53\% in Case 2 of Overall Qualitative evaluation methods with minimal GPU usage. Despite challenges such as near-synonymous subject labels and sparse domain terms, our system still secured the \textbf{ 9th} position in both quantitative (all subjects) and qualitative evaluations. These findings highlight the potential of cross-lingual pipelines and public code for fragmented training and inference, even under resource-constrained conditions.

\section{Task Description}
\label{sec:task}
The LLMs4Subjects shared task \cite{dsouza-EtAl:2025:SemEval2025} challenges participants to develop LLM-based systems for automated subject tagging in a national technical library. In this task, systems must assign one or more subject codes to each article, drawing from an extensive GND \cite{GND} subjects taxonomy. Participants are provided with a human-readable version of the taxonomy along with a bilingual data set (English / German) from TIBKAT \cite{TIBKAT} that includes various record types such as \texttt{article}, \texttt{book}, \texttt{conference}, \texttt{report}, and \texttt{thesis}.

\subsection{Dataset Description}
\label{sec:data}

\paragraph{}
The \textbf{LLMs4Subjects} corpus combines two complementary resources that are released together for SemEval 2025~Task 5:
\begin{enumerate}[label=\alph*)]
    \item \textbf{GND subject taxonomy:}  machine-readable export of more than 200000 controlled subject descriptors\footnote{`GND Sachbegriff’ in the German National Library authority file}.  
          Each entry provides a persistent identifier, the preferred German label, optional alternative labels, and explicit \textit{broader}, \textit{narrower}, and \textit{related} relations.  

    \item \textbf{TIBKAT technical records:}  123589 bibliographic records (title + abstract) drawn from the open access catalog of the German National Library of Science and Technology.  
          The corpus is bilingual (52 \% German, 48 \% English) and covers five record types: \textit{article}, \textit{book}, \textit{conference}, \textit{report}, and \textit{thesis}.  
          We follow the official split with 81937 documents for training, 13666 for development, and 27986 for blind testing; gold GND labels are provided for the first two splits only.
\end{enumerate}

\section{Proposed System}
\label{sec:method}

Our method has two well-defined stages.  
First, we create fixed embeddings for every data.
Second, we train a small neural module that makes the embeddings of the articles match the embeddings of the GND subjects.

All subject labels (originally in German) are translated into English.  
We then use pre-trained Sentence-Transformer models to encode both
subjects and articles.  
For articles, we join the title and abstract before encoding.  
Each subject and each article are now represented by a single
768-dimensional vector.  
These embeddings serve as input to our high-capacity transformation model, which is designed to enhance the semantic alignment between article embeddings and subject embeddings.

\begin{itemize}
\item \textbf{Dimension-as-Token Projection and Reshaping:}  
      We treat every number in the 768-long vector as its own
'token'.  
      A lightweight Burst-Attention layer with 16 hidden units and two heads lets these tokens exchange information.
\item \textbf{Feed-Forward MLP:}  
      The attention output passes through a three-layer feed-forward
      network with dropout.  
      The result is the final vector of aligned articles.
\end{itemize}
Together, these components transform initial embeddings into a
space where articles are more closely aligned with their relevant
subjects, thus improving the effectiveness of subject retrieval.

\paragraph{Training procedure.}
During training only article vectors go through the alignment module(transformation module).  
For each article, we also retrieve its gold subject vectors and average
them.  
The loss function moves the aligned article vector closer to this
average and further away from a few randomly chosen, incorrect subject
vectors.  
The subject vectors themselves remain fixed.

\paragraph{Inference procedure.}
At test time the alignment module is frozen.  
A new article is embedded, passed through the module, and compared using cosine similarity to the stored subject vectors.  
The \(k\) subjects with the smallest distance are returned as
recommendations.

\paragraph{Design motivation.}
Leaving the large pretrained transformer models untouched avoids costly
fine-tuning and keeps subject and article vectors consistent.  
The tiny alignment module is enough to correct systematic differences
between the two sets of vectors while adding very little computation.

\begin{figure}[!ht]
    \centering
    \includegraphics[scale=0.56]{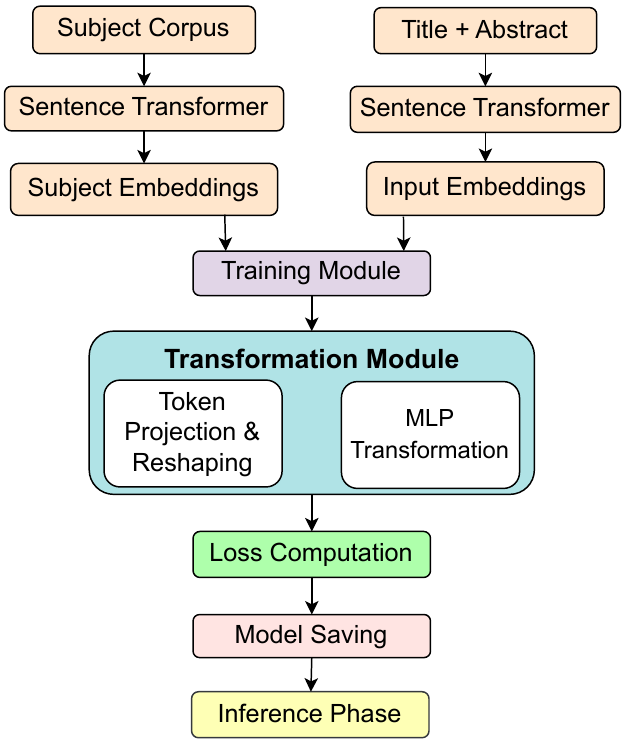}
    \caption{System Architecture. The model processes subject and article embeddings through a transformation module (Token Projection \& Reshaping + MLP) and optimizes via margin-based loss}
    \label{Fig:diagram1}
\end{figure}

Figure~\ref{Fig:diagram1} illustrates the complete system architecture.
It begins by extracting embeddings for both the subject corpus and
article texts, followed by a training module that applies the
transformation block and computes a margin-based loss.  After
training, the resulting model is saved for inference, where articles are
matched against subject embeddings to retrieve the most relevant
subjects.

\subsection{Data Preprocessing}
\label{sec:preprocessing}

In this stage, we convert all subject names and article texts into fixed-size vectors using off-the-shelf sentence-transformer models. These vectors form the inputs to our alignment module.

\paragraph{Subject corpus}  
We begin with the GND subject list. Each entry has a unique \textit{ code} and a German-language \textit{Name} (Subject Name). We translate every name into English using the Helsinki-NLP/opus-mt-de-en model \cite{tiedemann2020opus}. Next, we embed each translated name with a Sentence-Transformer. Concretely, for the \(i\)th subject name \(n_i\) we compute
\begin{equation}
     \mathbf{s}_i = ST(n_i) \in \mathbb{R}^{d}, \quad d = 768.
\end{equation}
  
Collecting all \(N\) subject vectors yields the subject matrix
\begin{equation}
       \mathbf{S}
   = 
   \begin{bmatrix}
     \mathbf{s}_{1}\\
     \mathbf{s}_{2}\\
     \vdots\\
     \mathbf{s}_{N}
   \end{bmatrix}
   \in \mathbb{R}^{N \times d}.
\end{equation}

\paragraph{Article corpus} 
For the training data, each sample comprises a \textit{title}, an \textit{abstract}, and a set of gold-standard (correct) subject labels. We concatenate the title $t_j$ and abstract $a_j$ into a single text $x_j$, and compute its embedding as
\begin{equation}
    \mathbf{m}_j = ST(x_j) \in \mathbb{R}^d.
\end{equation}
For each training sample $j$, every gold subject name $s_{jl}$ (with $l=1,2,\ldots,k_j$) is embedded, and the resulting embeddings are averaged to obtain a single representative embedding:
\begin{equation}
    \mathbf{g}_j = \frac{1}{k_j} \sum_{l=1}^{k_j} ST(s_{jl}) \in \mathbb{R}^d.
\end{equation}
Thus, the training document embeddings and the corresponding gold subject embeddings are aggregated into the matrices:
\begin{equation}
    \mathbf{M} = \begin{bmatrix} \mathbf{m}_1 \\ \mathbf{m}_2 \\ \vdots \\ \mathbf{m}_M \end{bmatrix} \in \mathbb{R}^{M \times d}, \quad \mathbf{G} = \begin{bmatrix} \mathbf{g}_1 \\ \mathbf{g}_2 \\ \vdots \\ \mathbf{g}_M \end{bmatrix} \in \mathbb{R}^{M \times d}.
\end{equation}

\paragraph{Embedding models}  
For English text we use \textbf{all-mpnet-base-v2} \cite{reimers2019sentence}.  
For German text we use \textbf{T-Systems-onsite/german-roberta-sentence-transformer-v2} \cite{t-systems-german-roberta-sentence-transformer-v2}. Both models produce 768-dimensional outputs.

\subsection{Dimension-as-Token Projection and Reshaping}
To enhance expressiveness, we reinterpret each dimension of an article embedding $\mathbf{m}$ as an individual "token". Starting with an embedding $\mathbf{m}$ of shape (batch\_size $B$, $d$), we first reshape it to a tensor of shape ($d$, $B$, 1). This operation enables us to treat each scalar dimension as a separate token, which is then projected to a hidden representation of size 16 (denoted as \textit{model\_dim}). Next, we apply a Burst Attention \cite{sun2024burstattention} encoder  across these $d$ tokens to capture their interdependencies. Finally, the tokens are projected back to a single scalar and reshaped to recover an embedding of the original dimension $d$. The forward pass is expressed as:
    \begin{equation}
    \begin{aligned}
        \mathbf{\Tilde{x}} &= \text{Reshape}(\mathbf{m}) \in \mathbb{R}^{d \times B \times 1}, \\
        \mathbf{X}_0 &= \text{Linear}_{\text{in}}(\mathbf{\Tilde{x}}) \in \mathbb{R}^{d \times B \times \text{model\_dim}}, \\
        \mathbf{X}_L &= \text{BurstAttnEnc}(\mathbf{X}_0), \\
        \mathbf{X}_{\text{out}} &= \text{Linear}_{\text{out}}(\mathbf{X}_L) \in \mathbb{R}^{d \times 1}, \\
        \mathbf{x}_{\text{attn}} &= \text{Reshape}^{-1}(\mathbf{X}_{\text{out}}) \in \mathbb{R}^d.
    \end{aligned}
    \end{equation}

Within the \textit{BurstAttnEnc}, a Burst Attention\cite{sun2024burstattention} layer is used. Initially, layer normalization is applied to the input, after which multi-head self-attention is computed using 2 attention heads, each with a per-head dimension of 8. A dropout of 0.03 is applied to the attention output before it is passed to a feedforward network. This network expands the hidden representation to 64 dimensions, applies a ReLU activation, and includes another dropout of 0.03, with residual connections added throughout to ensure stable learning. The Burst Attention encoder is defined as:
\begin{equation}
\begin{aligned}
    \mathbf{X}^{(0)} &= \mathbf{X}_0, \\
    \mathbf{Z}^{(l-1)} &= \text{LayerNorm}(\mathbf{X}^{(l-1)}), \\
    \mathbf{\Tilde{X}}_{\text{attn}}^{(l)} &= \text{MultiHead}(\mathbf{Z}^{(l-1)}),  \\
    \mathbf{Y}^{(l)} &= \mathbf{X}^{(l-1)} + \text{Dropout}(\mathbf{\Tilde{X}}_{\text{attn}}^{(l)}), \\
    \mathbf{U}^{(l)} &= \text{LayerNorm}(\mathbf{Y}^{(l)}), \\
    \mathbf{\Tilde{X}}_{\text{ff}}^{(l)} &= W_2 \cdot \text{ReLU}(W_1 \cdot \mathbf{U}^{(l)} + b_1) + b_2,  \\
    \mathbf{X}^{(l)} &= \mathbf{Y}^{(l)} + \text{Dropout}(\mathbf{\Tilde{X}}_{\text{ff}}^{(l)}).
\end{aligned}
\end{equation}
Here, the weight matrices $W_1 \in \mathbb{R}^{16 \times 64}$ and $W_2 \in \mathbb{R}^{64 \times 16}$ are learnable parameters within the feed-forward sub-module.

\subsection{Feed-Forward MLP}
Subsequent to obtaining the intermediate representation $\mathbf{x}_{\text{attn}}$ from the burst-attention block, we further refine the embedding using a Feed-Forward Multi-Layer Perceptron (MLP). This MLP is composed of three linear layers with ReLU activations and incorporates a dropout of 0.03 at each stage to reduce the risk of overfitting. The purpose of the MLP is to transform the intermediate representation back into the original embedding space, thus producing the final transformed embedding $\mathbf{z}$. This process is mathematically described as:
\begin{equation}
    \begin{aligned}
        \mathbf{h}_1 &= \text{ReLU(Dropout(}W_1^{\text{MLP}} \mathbf{x}_{\text{attn}} + b_1^{\text{MLP}})), \\
        \mathbf{h}_2 &= \text{ReLU(Dropout(}W_2^{\text{MLP}} \mathbf{h}_1 + b_2^{\text{MLP}})), \\
        \mathbf{z} &= W_3^{\text{MLP}} \mathbf{h}_2 + b_3^{\text{MLP}}.
    \end{aligned}
\end{equation}
In this formulation, $\mathbf{h}_1$ and $\mathbf{h}_2$ represent the outputs of the first and second hidden layers, respectively, and $\mathbf{z}$ is the output of the final layer. The weight matrices $W_1^{\text{MLP}} \in \mathbb{R}^{d \times 256}$, $W_2^{\text{MLP}} \in \mathbb{R}^{256 \times 256}$, and $W_3^{\text{MLP}} \in \mathbb{R}^{256 \times d}$, along with their corresponding biases $b_1^{\text{MLP}}$, $b_2^{\text{MLP}}$, and $b_3^{\text{MLP}}$, are learned during training.

Finally, the overall transformation function, denoted as \(\mathrm{transform}()\), is defined as the composition of the above operations, and for an input $\mathbf{m}$ it is computed as:
\vspace{-15pt}
\begin{equation}
\begin{split}
    \mathrm{transform}(\mathbf{m}) &= \text{MLP} \Big( \text{BurstAttnEnc} \big( \\
    &\quad \text{Linear}_{\text{in}}(\mathrm{Reshape}(\mathbf{m})) \big) \Big),
\end{split}
\end{equation}
\vspace{-1pt}
which outputs \(\mathbf{z} \in \mathbb{R}^d\). This architecture is designed to capture the complex interdimensional relationships present in the initial embeddings, ultimately yielding a more robust and semantically aligned representation.

\section{Training and Inference}
\label{sec:training}

During training, we learn only the small transformation block, keeping the Sentence-Transformer weights and the subject matrix \(\mathbf{S}\) fixed.  For each article \(j\), with its raw embedding $\mathbf{m}_j$ we apply the learned transform to obtain the anchor embedding
   $\mathbf{a}_j \;=\; \mathrm{transform}(\mathbf{m}_j)\;\in\;\mathbb{R}^d.$
Let the averaged gold‐subject embedding be \(\mathbf{p}_j\), and let \(\mathbf{n}_{j1},\dots,\mathbf{n}_{jk}\) denote \(k\) negative subject embeddings sampled from \(\mathbf{S}\).  Using cosine distance 
$d(\mathbf{x},\mathbf{y})
   = 1 \;-\; \frac{\mathbf{x}\!\cdot\!\mathbf{y}}{\|\mathbf{x}\|\;\|\mathbf{y}\|}$ ,
we define the margin-based loss for article \(j\) as
\begin{equation}
    \mathcal{L}_j 
    = \sum_{i=1}^k 
        \max\Bigl\{0,\;\alpha 
          +\,d(\mathbf{a}_j,\mathbf{p}_j)
          -\,d(\mathbf{a}_j,\mathbf{n}_{ji})\Bigr\},
    \quad 
\end{equation}
$\alpha=0.2,$
Minimizing \(\mathcal{L}_j\) brings the transformed article closer to its correct subjects while pushing it away from the incorrect ones.

We train for 20 epochs with batch size 4, using the AdamW optimizer and a cosine annealing scheduler.  In each batch, only the embeddings of the article, their positives and the sampled negatives are passed through \(\mathrm{transform}(\cdot)\); the embeddings of the subject remain frozen.

\paragraph{Inference}  
At inference time, we first transform and cache all GND subject embeddings as  
$\mathbf{S}' = \mathrm{transform}(\mathbf{S}),$
so that each row \(\mathbf{s}'\) is the aligned vector for one subject. For a new article (title and abstract), we compute its embedding  
\(\mathbf{m} = ST(\text{title}\|\text{abstract})\)  
using a Sentence-Transformer, then apply the trained mapping to obtain  
\(\mathbf{a} = \mathrm{transform}(\mathbf{m})\). Next, we measure the cosine-distance between \(\mathbf{a}\) and each precomputed subject vector \(\mathbf{s}'\):  
$d(\mathbf{a},\mathbf{s}') = 1 - \frac{\mathbf{a}\cdot\mathbf{s}'}{\|\mathbf{a}\|\;\|\mathbf{s}'\|}\quad\forall\,\mathbf{s}'\in\mathbf{S}'.$
Finally, we sort all subjects by increasing distance and return the \(k\) codes with the smallest values.

\section{Experimental Setup}

Our experiments were carried out on a bilingual data set partitioned into training, development and test sets for both English and German articles. The training set was used for margin-based optimization, the development set for intermediate performance monitoring, and the test set was reserved for official SemEval evaluation. 

The embedding dimension is set to 768, and our model is configured with a hidden dimension (\texttt{model\_dim}) of 16, 2 attention heads, a single Burst Attention layer, an MLP hidden dimension of 256, and a dropout rate of 0.03. The training employs a batch size of 4 and utilizes 15 negative samples per anchor-positive pair over 20 epochs.

We apply a margin-based loss with a margin of 0.2, optimizing the model using the AdamW optimizer with a learning rate of $1\times10^{-4}$ (without weight decay) and a cosine annealing learning rate scheduler ($T_{max} = 20$, $\eta_{\min} = 1\times10^{-6}$). All experiments were executed on Google Colab L4 GPUs under Python 3.9, using PyTorch 1.13.1, SentenceTransformers 2.2.2, pandas 1.5.3, numpy 1.23.5, and tqdm 4.64.1. Evaluation measures included Precision@k, Recall@k, and F1@k across multiple cutoffs, along with average recall, as reported in Section~\ref{sec:results}.

% \subsection{Implementation Details}
% We use \texttt{PyTorch} for both English and German training:
% \begin{itemize}
%     \item \(\text{model\_dim}=16\), 2 attention heads, 1 layer of BurstAttention.
%     \item \(\text{mlp\_hidden\_dim}=256\), dropout $=0.03$.
%     \item \(\text{batch\_size}=4, \text{lr}=1\times10^{-4}, \text{weight\_decay}=0\).
%     \item Negative sampling $k=5$ or $15$ depending on GPU memory.
%     \item 5--20 training epochs with a cosine LR schedule.
% \end{itemize}

\section{Results}
\label{sec:results}

We report the performance of our system on the official test segment as well as on two qualitative evaluation subsets. For evaluation, we use Precision @ k (P @ k), Recall @ k (R @ k), and F1 @ k. Precision @ k is defined as the number of relevant items in the top \(k\) predictions divided by  Total number of items in \(k\),
Recall @ k is defined as the number of relevant items in the top \(k\) predictions divided by the total number of relevant items.
The F\(\beta\)-score is computed as 
\begin{equation}
   F_{\beta} = \frac{(1+\beta^2) \times \text{Precision @ k} \times \text{Recall @ k}}{\beta^2 \times \text{Precision @ k} + \text{Recall @ k}} 
\end{equation}
and in our experiments we set \(\beta=1\) to obtain the F1@k score.

\paragraph{Quantitative Results} 
Evaluation~1 of the shared task compared systems in a fully automatic setting in the blind TIBKAT test corpus.  
Participants submitted the \emph{top–50} GND subject codes for every English or German record, and the organisers calculated P @ k, R @ k, and F1 @ k for each \(k\in\{5,10,\dots,50\}\).  
Scores were released at three granularities: (i) language level (en vs.\ de), (ii) record–type level (five technical record types), and (iii) the combined language–record level.  
Systems were ranked by the \emph{average Recall @ k} over all cut-offs, a criterion in which our \textbf{NBF} run placed \textbf{9th}.

\paragraph{Qualitative Results} 
Evaluation~2 complemented the automatic leaderboard with a manual assessment by subject specialists from TIB.  
A stratified sample of 140 records—ten from each of fourteen discipline codes (arc, che, elt, fer, his, inf, lin, lit, mat, oek, phy, sow, tec, ver)—was drawn.  
For every record, librarians inspected the \emph{top–20} subjects proposed by each system and labelled them \texttt{Y} (correct), \texttt{I} (irrelevant but technically admissible), or \texttt{N}/blank (incorrect).  
Two leaderboards were produced: \textbf{Case 1}, counting both \texttt{Y} and \texttt{I} as correct, and \textbf{Case 2}, counting only \texttt{Y}.  
These settings correspond exactly to Tables~\ref{tab:nbsCase1} and~\ref{tab:nbsCase2}. 

Table~\ref{tab:nbsAll} summarizes the overall quantitative results on the all-subjects evaluation (official test set) at cutoffs \(k \in \{5,10,15,\dots,50\}\). The model achieves an average recall of 32.24\% across these cutoffs, although the precision remains relatively low. Based on these official metrics, our system ranked \textbf{9th} in the quantitative track.

To better understand our design choices, we conducted experiments on two different subsets of the test data. Table~\ref{tab:nbsCase1} shows the overall qualitative results for Case~1, a subset characterized by common and less ambiguous subject labels. On this subset, the model achieves higher Precision, Recall, and F1 (with averages of 39.54\%, 43.16\%, and 39.03\%, respectively). In contrast, Table~\ref{tab:nbsCase2} presents overall Qualitative results for Case 2, which consists of articles with more specialized or challenging subject mappings, yielding lower average metrics (Precision = 19.00\%, Recall = 31.53\%, F1 = 22.40\%). These analyses demonstrate that while our parameter-efficient approach is competitive overall, its performance is sensitive to the complexity of subject labels.

\begin{table}[!tb]
\centering
\small
\caption{All-subjects (Overall Quantitative) results on the official test set for Team NBF (best run). }
\label{tab:nbsAll}
\begin{tabular}{lccc}
\toprule
\textbf{k} & \textbf{P@k} & \textbf{R@k} & \textbf{F1@k} \\
\midrule
5   & 0.0835 & 0.1699 & 0.1120 \\
10  & 0.0594 & 0.2329 & 0.0946 \\
15  & 0.0475 & 0.2742 & 0.0809 \\
20  & 0.0401 & 0.3048 & 0.0708 \\
25  & 0.0351 & 0.3305 & 0.0635 \\
30  & 0.0314 & 0.3515 & 0.0576 \\
35  & 0.0285 & 0.3694 & 0.0529 \\
40  & 0.0261 & 0.3839 & 0.0489 \\
45  & 0.0241 & 0.3974 & 0.0455 \\
50  & 0.0225 & 0.4095 & 0.0426 \\
\midrule
\multicolumn{4}{l}{\textbf{Average Recall} = \textbf{0.3224}} \\
\bottomrule
\end{tabular}
\end{table}

\begin{table}[!tb]
\centering
\small
\caption{Case 1 Results (Overall Qualitative Evaluation: Common Subjects).}
\label{tab:nbsCase1}
\begin{tabular}{lccc}
\toprule
\textbf{k} & \textbf{P@k} & \textbf{R@k} & \textbf{F1@k} \\
\midrule
5   & 0.4405 & 0.2068 & 0.2815 \\
10  & 0.4099 & 0.3738 & 0.3910 \\
15  & 0.3753 & 0.5097 & 0.4323 \\
20  & 0.3559 & 0.6362 & 0.4565 \\
\midrule
\textbf{Average} & \textbf{0.3954} & \textbf{0.4316} & \textbf{0.3903} \\
\bottomrule
\end{tabular}
\end{table}

\begin{table}[!tb]
\centering
\small
\caption{Case 2 Results (Overall Qualitative Evaluation: Specialized Subjects).}
\label{tab:nbsCase2}
\begin{tabular}{lccc}
\toprule
\textbf{k} & \textbf{P@k} & \textbf{R@k} & \textbf{F1@k} \\
\midrule
5   & 0.2344 & 0.1718 & 0.1983 \\
10  & 0.1956 & 0.2830 & 0.2313 \\
15  & 0.1734 & 0.3666 & 0.2354 \\
20  & 0.1565 & 0.4398 & 0.2309 \\
\midrule
\textbf{Average} & \textbf{0.1900} & \textbf{0.3153} & \textbf{0.2240} \\
\bottomrule
\end{tabular}
\end{table}

\subsection{Error Analysis}
Although our system shows potential in assigning subject labels, we found two main sources of error. First, the system struggles with synonym overlaps. For instance, near-synonymous labels such as \emph{``Natural Language Processing''} and \emph{``Computational Linguistics''} often appear together in the training data without sufficient distinction, causing the model to confuse these overlapping concepts and misclassify the subjects. 

Second, errors frequently occur with sparse domain terms. Articles addressing highly specialized or infrequent topics tend to be mislabeled because the negative sampling does not adequately cover these less common domains. These challenges suggest that future work should focus on enhancing the model's ability to differentiate between similar subject labels and on improving the representation of niche topics in the training process.

\section{Conclusion and Future Work}
We presented our system, which employs a novel dimension-as-token self-attention mechanism (\emph{Burst Attention}) on top of Sentence Transformers. Our experiments demonstrate that, even with an ultra-light hidden dimension (16) and a single attention layer, the approach is competitive in terms of recall, achieving average recall rates of 32.24\%, 43.16\% and 31.53\% in the Overall Quantitative, Qualitative Case 1, and Qualitative Case 2 evaluations, respectively. These results indicate that our model is effective at capturing relevant subject information under resource constraints, although there is still room for improvement.

In future work, we plan to explore deeper stacking of Burst Attention layers, more advanced negative sampling strategies, and the incorporation of hierarchical subject ontologies for improved synonym disambiguation. Additionally, we intend to investigate the integration of multimodal features to further enhance system performance. These enhancements aim to further improve both the accuracy and robustness of our subject classification system.

\section*{Acknowledgments}
We thank the organizers of the SemEval-2025 Task 5 \cite{dsouza-EtAl:2025:SemEval2025} for the data and evaluation framework, as well as the PyTorch, Hugging Face and SentenceTransformers communities for their invaluable open-source tools.

\end{document}